%% file: ijcai25.tex
\documentclass{article}
\usepackage{ijcai25}

\usepackage{times}
\usepackage{soul}
\usepackage{url}
\usepackage[hidelinks]{hyperref}
\usepackage[utf8]{inputenc}
\usepackage[small]{caption}
\usepackage{graphicx}
\usepackage{amssymb}
\usepackage{amsmath}
\usepackage{amsthm}
\usepackage{booktabs}
\usepackage{algorithm}
\usepackage{algorithmic}
\usepackage{xspace}
\usepackage[switch]{lineno}

\usepackage{multirow}

\title{LEKA: LLM-Enhanced Knowledge Augmentation}

\author{
Xinhao Zhang$^1$
\and
Jinghan Zhang$^1$
\and
Fengran Mo$^2$
\and
Dongjie Wang$^3$
\and
Yanjie Fu$^4$
\And 
Kunpeng Liu$^{1}$\thanks{Corresponding author.}\\ 
\affiliations
$^1$Portland State University, USA\\
$^2$University of Montreal, Canada\\
$^3$University of Kansas, USA\\
$^4$Arizona State University, USA\\ 
\emails
\{xinhaoz, jinghanz, kunpeng\}@pdx.edu,
fengran.mo@umontreal.ca,
wangdongjie@ku.edu,
yanjie.fu@asu.edu 
}

\newcommand{\model}{LEKA\xspace}
\newcommand{\modelbf}{\textbf{LEKA}\xspace}

\begin{document}

\maketitle

\begin{abstract}
    Humans excel in analogical learning and knowledge transfer and, more importantly, possess a unique understanding of identifying appropriate sources of knowledge. From a model's perspective, this presents a unique challenge. If models could autonomously retrieve knowledge relevant for transfer or decision-making to solve problems, they would transition from passively acquiring to actively accessing and learning from knowledge. However, filling models with knowledge is relatively straightforward—it simply requires more training and accessible knowledge bases. The more complex task is teaching models about which knowledge can be analogized and transferred. Therefore, we design a knowledge augmentation method, \model, for knowledge transfer that actively searches for suitable knowledge sources that can enrich the target domain's knowledge. This \model method extracts key information from the target domain's textual information, retrieves pertinent data from external data libraries, and harmonizes retrieved data with the target domain data in feature space and marginal probability measures. We validate the effectiveness of our approach through extensive experiments across various domains and demonstrate significant improvements over traditional methods in automating data alignment and optimizing transfer learning outcomes.
\end{abstract}

\input{main/1-intro}

\input{main/2-related}

\input{main/3-method}

\input{main/4-exp}
\input{main/5-concl}

\newpage
\input{main/7-acknowledgment}
\bibliographystyle{named}
\bibliography{ijcai25}

\end{document}

%% file: main/1-intro.tex
\section{Introduction}
Humans are good at identifying relevant sources of knowledge. This is an ability that is rooted in our capability for analogical reasoning and knowledge management. In contrast, artificial intelligence models do not inherently possess this intuition: they require explicit instructions and systematic training to identify and utilize relevant information. This gap presents a challenge in enriching domain knowledge and enhancing data augmentation.

Knowledge augmentation is vital for improving model performance, especially in domains with limited information or complex data structures~\cite{tang2020onlineaugment}. Knowledge transfer is a crucial method within knowledge augmentation for improving learning performance by transferring knowledge from external information sources~\cite{khodaee2024knowledge}. In data-limited scenarios, effective knowledge sourcing can bridge domain information gaps and enhance model robustness using relevant external information. In these scenarios, knowledge augmentation, especially knowledge transfer, can reduce reliance on extensive target domain data by strategically selecting sources matching target needs.

Despite its potential, traditional knowledge augmentation methods often involve manual intervention~\cite{ringwald2021adaptiope,nam2024tabular,zhang2024dynamic}, where human experts select source domains based on their subjective interpretation of the target domain's requirements. Using tabular learning as an example highlights several challenges: (1) structural and format differences across domains hinder data alignment and integration; (2) discrepancies in tasks and content between the selected source and target domains can diminish the success of knowledge augmentation; and (3) extensive data preprocessing is often required to properly match the chosen source domain dataset. Additionally, relying on human expertise for source domain selection can lead to biases and inefficiencies, as these choices are typically made based on prior knowledge rather than a rigorous, data-driven analysis.

A natural idea is to construct an automated search for datasets with relevant knowledge in a database that has a similar structure to the target data, such as utilizing Retrieval-Augmented Generation (RAG)~\cite{gao2023retrieval,lewis2020retrieval}. RAG combines retrieval systems with generative language models, and it enhances the model's capabilities by providing access to an external library~\cite{guo2017deepfm,zhang2025ratt,SHI201881}. For dataset retrieval, some existing works~\cite{fleischer2024rag,siriwardhana2023improving} use the retrieved documents as contextual information for generation and significantly enrich the model's knowledge base.

However, this direct method is often impractical due to the high costs of embedding entire domains or datasets~\cite{seemakhupt2024edgerag,jin2024ragcache}. Additionally, retrieved knowledge may not align effectively with the intended application~\cite{edge2024local,li2025oreo}. Such misalignments typically result from variations in data distribution, feature spaces, or contextual differences between source and target data. Furthermore, retrieved data quality, like noise or incomplete features, can also hinder its effectiveness for augmenting data via knowledge augmentation. Thus, in such situations, identifying the optimal knowledge source providing the most relevant, high-quality data tailored to specific target data needs via LLM capabilities remains a substantial challenge.

\begin{figure*}
    \centering
    \includegraphics[width=0.95\textwidth]{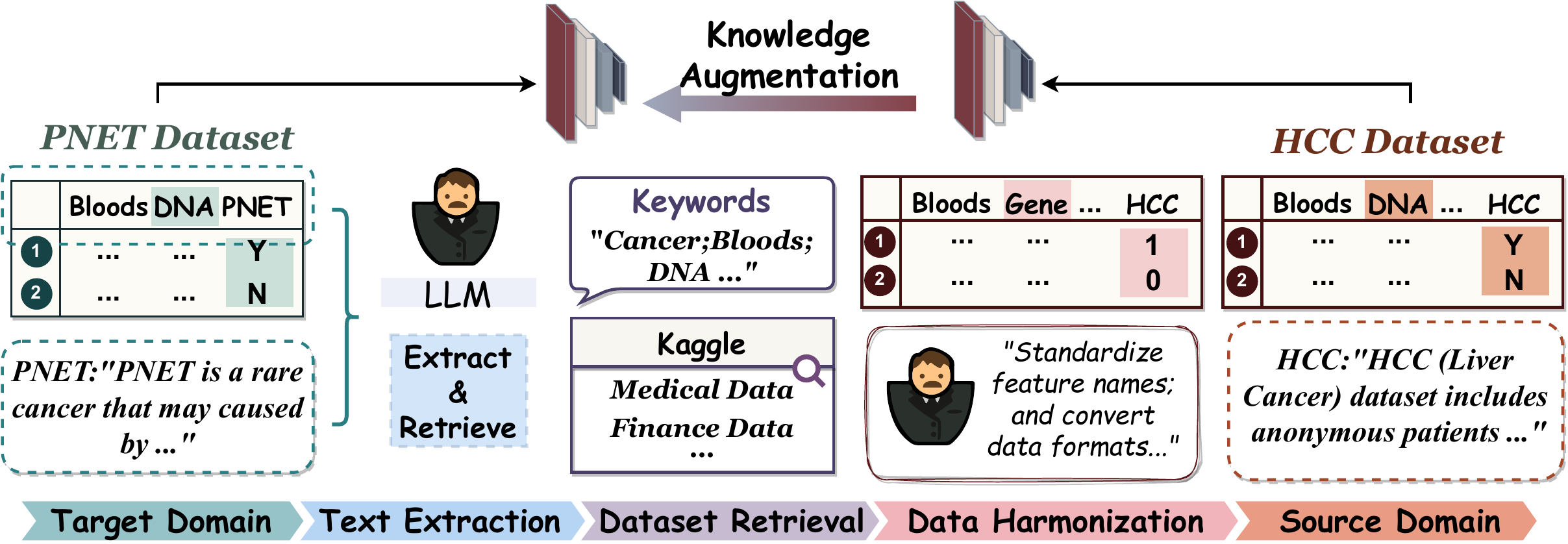}
    \caption{Example of \model. We adopt an LLM to retrieve proper source domain data to transfer knowledge to a data-limited target domain. The LLM extracts the key information of the target data to retrieve a relevant dataset; then, we adopt the LLM for harmonization.}
    \label{fig:intro}
\end{figure*}

\paragraph{Our Targets.} We aim to address three main challenges in retrieving source domain data: (1) how to effectively extract key information from the target domain with limited computational cost; (2) how to design an efficient, automatic source domain retrieval for knowledge augmentation; and (3) how to automatically harmonize the retrieved source data with the target domain to optimize transfer efficiency and improve learning outcomes.

\paragraph{Our Approach.} To address the challenges above, we design \textbf{L}LM-\textbf{E}nhanced \textbf{K}nowledge \textbf{A}ugmentation (\modelbf), a novel and automated data retrieval and augmentation method by knowledge augmentation. Specifically, (1) we utilize an LLM to extract the key textual information of the target domain; (2) we deploy dataset-RAG in an external database to efficiently extract knowledge relevant to the target domain dataset. The RAG libraries can contain large amounts of continuously updated datasets, ensuring that the data retrieved from these libraries is more precise and timely; (3) then the LLM automatically harmonizes the retrieved source data with the target domain in feature space and marginal probability measures to enhance downstream task learning performance. The data harmonization reduces the structural and semantic differences between the source and target data. This data harmonization is crucial for reducing domain shift, which optimizes transfer efficiency and improves the overall learning outcomes of the model.

As shown in Figure~\ref{fig:intro}, consider a target domain focused on predicting a rare cancer, so direct learning is challenging due to insufficient sample information. To augment the data with external information would typically require human experts with knowledge of both cancer physiological indicators and machine learning to retrieve datasets. In contrast, our method can automate the retrieval of a common cancer dataset as the knowledge source with extensive feature information. This retrieved dataset possesses a more complete and transparent feature space with a structure similar to the target domain. The LLM then adjusts the retrieved data based on the characteristics of the target domain. In this way, we achieve a highly harmonized source dataset for knowledge augmentation.

In summary, our contribution includes:
\begin{enumerate}
    \item We introduce a novel automated data augmentation method, \model, which utilizes an automated retrieval method for external data and harmonizes it with target data for knowledge augmentation.
    \item We develop a novel paradigm by incorporating an LLM into the data harmonization process to optimize data space and structure to enhance downstream machine learning performance.
    \item We conduct a series of experiments to validate the effectiveness and robustness of our \model method across different tasks. Experimental results demonstrate that our method has clear advantages over existing methods.
\end{enumerate}

%% file: main/2-related.tex
\section{Related Work}
\subsection{Knowledge Transfer}
Knowledge transfer is a knowledge augmentation method that improves learning on new tasks by transferring knowledge from a related task~\cite{alyafeai2020survey,wang2022semi,wang2025diversity}. Knowledge transfer allows cross-domain knowledge transformation despite data distribution or feature space differences. Specifically, knowledge transfer adapts models developed for one task to perform better on different but related tasks, as it adjusts the feature mappings and decision boundaries to suit new knowledge~\cite{han2021robust,yordanov2021few}. 
Finding suitable pre-training knowledge is crucial for knowledge transfer because it significantly enhances the model's effectiveness by providing a relevant starting point. While many methods exist to adjust existing knowledge within the same or across different domains to improve transfer learning outcomes, the process of retrieving certain knowledge still heavily relies on manual effort.
\subsection{Data Harmonizing with LLMs}

Data harmonization applying LLMs offers significant benefits by leveraging their natural language capabilities to standardize and integrate diverse datasets and further enhance model performance through improved data consistency~\cite{cao2009agent,zhang2024prototypical,liu2021automated,durante2024agent}. However, this approach has several challenges, including the high computational costs of training LLMs and potential biases in training data, which can adversely affect the harmonization process~\cite{feng2021survey,xie2024scoring,zhang2025blindspotnavigationllm}. Furthermore, the risk of overfitting remains a concern, as models may become overly specialized in training data nuances, reducing their effectiveness on new datasets.

\subsection{Retrieval Augmented Generation for Knowledge Augmentation}

Retrieval Augmented Generation (RAG)~\cite{lewis2020retrieval} for knowledge augmentation is a method that enhances the capabilities of generative language models by integrating information retrieval with model generation~\cite{hu2024rag}. This technique aids LLMs in tasks that demand deep and specific domain knowledge~\cite{zhang2024raft,huang2024survey}. RAG for knowledge augmentation provides access to expansive external libraries (collections of documents or domain-relevant knowledge), making it especially effective for transferring knowledge across different domains~\cite{siriwardhana2023improving}. RAG for knowledge augmentation embeds extensive databases directly into the generative process so that specific, domain-related information is both accessible and effectively utilized. In this way, it can significantly enhance the performance for complex tasks in knowledge augmentation scenarios.

\section{Preliminary}

\subsection{Definition}
\paragraph{{Definition 1.} (Domain)} A domain $\mathcal{D}$ is an ordered pair consisting of a feature space $\mathcal{X}$ and a marginal probability measure $P$ defined on this feature space. In other words, $\mathcal{D} = (X, P)$, where $\mathbf{X}= \{\mathbf{x} | \mathbf{x}_i \in X, i = 1, \ldots, n\}$ is an instance set. And $P$ is a probability measure that describes the probability of occurrence of feature vectors $\mathbf{x} \in \mathbf{X}$. This probability measure $P$ makes $(\mathcal{X}, \mathcal{B}(\mathcal{X}), P)$ a probability space, where $\mathcal{B}(\mathcal{X})$ is the Borel $\sigma$-algebra on $\mathcal{X}$.

\textit{\textbf{Source Domain.}} The source domain $\mathcal{D}_S$ consists of instances paired with labels $y$, represented as $\mathcal{D}_S = \{(\mathbf{x}, \mathbf{y}) | \mathbf{x}_i \in \mathcal{X}_S, y_i \in \mathcal{Y}_S, i = 1, \ldots, n^{S}\}$. Here, $\mathcal{X}_S$ is the feature space and $\mathcal{Y}_S$ is the label space for the source domain. This domain provides labeled data used to train models in preparation for transfer learning tasks.

\textit{\textbf{Target Domain.}} The target domain $\mathcal{D}_T$ typically contains a mix of unlabeled instances and a smaller set of labeled instances, denoted as $\mathcal{D}_T = \{\mathbf{x} \in \mathcal{X}_T\} \cup \{(\mathbf{x}, \mathbf{y}) | (\mathbf{x}, \mathbf{y}) \in \mathcal{X}_T \times \mathcal{Y}_T\}$. Here, $\mathcal{X}_T$ is the feature space and $\mathcal{Y}_T$ is the label space for the target domain. We aim to evaluate and fine-tune the transfer learning models with data in the target domain.

\paragraph{{Definition 2.} (Task)} A task $\mathcal{T}$ consists of a label space $\mathcal{Y}$ and a decision function $f$, formally noted as $\mathcal{T} = (\mathcal{Y}, f)$, where $\mathcal{Y}$ is a metric space that contains all possible labels, and $f$ is a mapping from the feature space $\mathcal{X}$ to a set of conditional probability measures on the label space $\mathcal{Y}$.

\textit{\textbf{Source Task.}} The learning task of the source task $\mathcal{T}_S$ is typically represented as learning a target function $f_S: \mathcal{X}_S \to \mathcal{Y}_S$, where $\mathcal{Y}_S$ is the label space of the source task.

\textit{\textbf{Target Task.}} The learning task of the target task $\mathcal{T}_T$ is typically represented as learning a target function $f_T: \mathcal{X}_T \to \mathcal{Y}_T$, where $\mathcal{Y}_T$ is the label space of the target task.

\paragraph{{Definition 3.} (Knowledge Transfer)} Knowledge transfer is an augmentation method that adopts observations from source domains and tasks, denoted as $\{(\mathcal{D}_{S_i}, \mathcal{T}_{S_i}) | i = 1, \ldots, m^S\}$. Here $m^S \in \mathbb{N}^{+}$ represents the number of source domains and tasks. Similar observations from target domains and tasks are denoted as $\{(\mathcal{D}_{T_j}, \mathcal{T}_{T_j}) | j = 1, \ldots, m^T\}$, where $m^T \in \mathbb{N}^{+}$. The goal of knowledge transfer is to utilize the knowledge embedded in the source domains $\mathcal{D}_{S_i}$ to enhance the performance of the learned decision functions $f_{T_j}$ across the target domains $\mathcal{D}_{T_j}$, for $j = 1, \ldots, m^T$.

\subsection{Problem Formulation}
We formulate the task to enhance the performance of knowledge transfer through automated data retrieval and harmonization using an LLM. Concretely, we improve the performance of the learned decision function $f_{T_j}$ by reconstructing and refining the source domain $\mathcal{D}_S$. In this way, we better utilize the knowledge implied in $\mathcal{D}_S$ as we mitigate domain shifts and facilitate a more effective transfer of learned models. Concretely, our optimization objective is to retrieve and reconstruct a source domain $\mathcal{D}_{S}^*$:

\begin{equation}
    \mathcal{D}_{S}^*= \underset{\hat{\mathcal{D}_S}}{\text{argmax}} \; \mathbf{\mathcal{P}}_{\mathcal{D}_T}(f_{T_j}),
\end{equation}

\noindent where $\mathcal{P}$ is the performance indicator of $f_{T_j}$ and $\hat{\mathcal{D}_S}$ is a reconstructed source domain aligned with $\mathcal{D}_T$.

%% file: main/3-method.tex
\section{Methodology}

\begin{figure*}
    \centering
    \includegraphics[width=\textwidth]{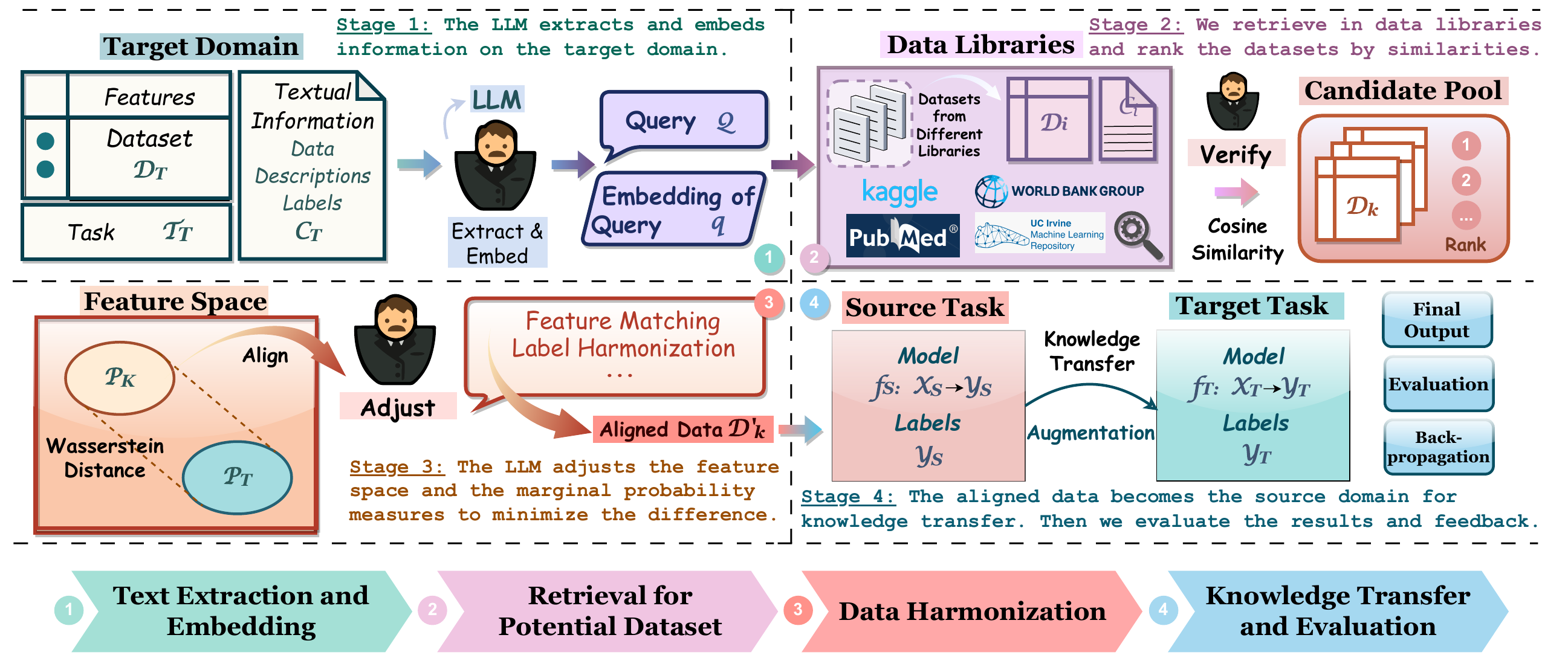}
    \caption{Framework of \model includes: 1) an LLM extracts and embeds the textual information of the target dataset, then 2) retrieves datasets in libraries, and 3) processes data harmonization. With harmonized datasets, we can transfer knowledge from the source dataset we construct to enhance learning on the target dataset.}
    \label{fig:method}
\end{figure*}
In this section, we introduce our novel knowledge augmentation method named \underline{\textbf{L}}LM-\underline{\textbf{E}}nhanced \underline{\textbf{K}}nowledge \underline{\textbf{A}}ugmentation (\textbf{LEKA}), designed to dynamically and automatically retrieve and refine source data to transfer knowledge and further enhance target data learning. Specifically, our study focuses on tabular datasets as the concrete form of data. The \model leverages an LLM to extract essential information from the target dataset, including data structure, feature names, and descriptions. Within this framework, the LLM uses its capability to analyze and synthesize data to optimize the selection and refinement of source datasets. It extracts and summarizes keywords for retrieval tailored to the target domain and task. After retrieval, the LLM refines and harmonizes these datasets in feature space and marginal probability measures with the target dataset. Figure~\ref{fig:method} illustrates the overview of the \model framework, which comprises three stages: (1) \textbf{Dataset Retrieval}; (2) \textbf{Data Harmonization}; and (3) \textbf{Knowledge Transfer and Evaluation}.


\subsection{Dataset Retrieval}

In this phase, our primary objective is to identify and retrieve a source dataset $\mathcal{D}_S$ similar to the target dataset $\mathcal{D}_T$ for effective pre-training in knowledge transfer scenarios. To achieve this, we utilize an LLM $\mathcal{A}$ to analyze and retrieve data that matches the structural properties and purposes of $\mathcal{D}_T$. A straightforward method is to embed the target dataset $\mathcal{D}_T$ by $\mathcal{A}$ to capture its essential features. However, this approach can be computationally intensive and prone to errors. Inspired by the strategies outlined in~\cite{zhang2024tifg}, we turn to leverage the textual information, including the data descriptions $\mathcal{C}_T$ and feature names $l_T$ of $\mathcal{D}_T$. These textual elements provide insights into the structure and purpose of $\mathcal{D}_T$, and the LLM can precisely focus on the semantic content of $\mathcal{D}_T$.

First, the LLM $\mathcal{A}$ constructs a query $\mathcal{Q}$ based on this textual information:

\begin{equation}
    \mathcal{Q} = \mathcal{A}(\mathcal{C}_T, l_T),
\end{equation}

\noindent and then embeds the query:
\begin{equation}
    q=\mathcal{A}(\text{embed}(\mathcal{Q})),
\end{equation}
\noindent where $\text{embed}(\cdot)$ is the embedding process that transforms the input into a vector representation. 

We then retrieve from a library $\mathcal{L}$ of datasets for the top-$k$ most relevant datasets $\{\mathcal{D}_k\}$ to query $\mathcal{Q}$. We evaluate the relevance of all datasets in $\mathcal{L}$ to the query $\mathcal{Q}$ by the cosine similarity of the embeddings and select the top-$k$ datasets with the highest similarity:

\begin{equation}
    \text{sim}(q, d_k) = \frac{q \cdot d_k}{\|q\| \|d_k\|} ,
\end{equation}
\noindent where $ d_k= \mathcal{A}(\text{embed}(\mathcal{D}_k)) $ is the embedding of a potential dataset $\mathcal{D}_k$'s textual information, and $\|\cdot\| \in \mathbb{R}$ is the norm function. 

In this phase, we map textual information to a high-dimensional vector space. Essentially, we approximate the probability distributions of the datasets in the library to retrieve $k$ source datasets most similar in structure and purpose to the target dataset. We leverage the geometric properties of vectors in high-dimensional space to assess and quantify dataset similarities. This prepares us for further dataset processing for transfer learning.

\subsection{Data Harmonization}

Now that we have $k$ potential source domain datasets, we turn to align these datasets with the target dataset $\mathcal{D}_T$ with an LLM. This alignment process adjusts the feature space and marginal probability measures of the source dataset $\mathcal{D}_k$ to closely match those of the target dataset $\mathcal{D}_T$. This process involves transformations of features and adjustments of distributions. Here we take $\mathcal{D}_k$ as an example.

\paragraph{Feature Space Transformation.}

We denote $\mathcal{X}_k$ and $\mathcal{X}_T$ as the feature spaces of the source dataset $\mathcal{D}_k$ and the target dataset $\mathcal{D}_T$, respectively. Our LLM constructs a mapping function $f: \mathcal{X}_k \rightarrow \mathcal{X}_T$ for alignment. This function transforms the features in $\mathcal{X}_k$ to minimize the distance between the transformed source features and the target features. Here, we aim to minimize the distance $ d(f(\mathcal{X}_k), \mathcal{X}_T) $, where $ d $ is a kernel distance:

\begin{equation}
    d(f(\mathcal{X}_k), \mathcal{X}_T) = \sqrt{\sum_{i,j} (\kappa(f(x_i^k), f(x_j^k)) - \kappa(x_i^T, x_j^T))^2}.
\end{equation}

\noindent Here, $\kappa(\cdot, \cdot) $ is a kernel function, $x_i^k$ and $x_j^k$ are the $i$-th and $j$-th feature vectors in the source dataset $\mathcal{D}_k$. We adopt a Gaussian kernel $\kappa(x, y) = \exp(-\gamma \| x - y \|^2) $, with $\gamma$ being a positive bandwidth parameter. This kernel function can effectively measure the similarity between points in high-dimensional spaces and captures both linear and non-linear relationships. Thus, it can handle the complexities inherent in high-dimensional data. Compared with Euclidean distances, kernel distances can capture the geometric structure of the data manifold. It provides a more robust and informative similarity measure for the source-target alignment. 

\paragraph{Marginal Probability Measures Harmonization.}
Our LLM aligns the marginal probability measures $\mathcal{P}_k$ and $\mathcal{P}_T$ of the source and target datasets. The LLM analyzes and refines textual information such as labels, feature names, and classification probability distributions. To measure the differences between $\mathcal{D}_k$ and $\mathcal{D}_T$, we adopt the Wasserstein distance:

\begin{equation}
    W(\mathcal{P}_k, \mathcal{P}_T) = \inf_{\gamma \in \Gamma(\mathcal{P}_k, \mathcal{P}_T)} \int_{\mathcal{X} \times \mathcal{X}} \|x - y\| \, d\gamma(x, y),
\end{equation}
\begin{equation}
    \mathcal{D}_k^{'} = \mathcal{A}(\mathcal{D}_k, \mathcal{P}_k, \mathcal{P}_T),
\end{equation}
where $ \Gamma(\mathcal{P}_k, \mathcal{P}_T)$ represents the set of all joint distributions with marginals $\mathcal{P}_k$ and $\mathcal{P}_T$ on $ \mathcal{X} \times \mathcal{X} $. The Wasserstein distance minimizes the transportation cost while preserving the geometric structure of the data distributions. Specifically, the Wasserstein distance calculates the minimum ``geographical'' cost required to move data from one distribution to another, where ``geographical'' cost refers to the cost of moving data from one point to another in the feature space.

Moreover, the Wasserstein distance is advantageous when dealing with distributions whose support sets (i.e., the effective range or set of the distributions) do not fully overlap. In real-world data applications, it is common for the source and target datasets to originate from different distributions. Their data points do not align perfectly and cover the same areas. In these cases, traditional distance metrics like the Euclidean distance poorly reflect the actual differences between the two datasets, as they merely measure differences in position but ignore the overall structure of the data distributions.

With this source-target alignment process, the LLM improves the efficiency of knowledge transfer from the source domain to the target domain. Thus, it enhances the overall performance of our transfer learning models. Further, we optimize data handling and boost the adaptability and accuracy of transfer learning for adoption in various applications.

\subsection{Knowledge Transfer and Evaluation}

After aligning the source dataset $\mathcal{D}_k'$ with the target dataset $\mathcal{D}_T$, and adjusting the marginal probability measures $\mathcal{P}_k'$ to $\mathcal{P}_T$, we proceed to integrate these aligned datasets into the transfer learning model. At this stage, our goal is to transfer knowledge from $\mathcal{D}_k'$ to $\mathcal{D}_T$ by minimizing a defined loss function while adapting the model to the target domain.

To achieve this, the transfer learning process focuses on updating the decision function $f_{T_j}$. Specifically, we aim to minimize the expected loss over the target dataset while incorporating the knowledge transferred from the source dataset:
\begin{equation}
    f_{T_j}^{*} = \arg\min_{f_{T_j}} \mathbb{E}_{(x, y) \in \mathcal{D}_T} [\mathcal{L}(f_{T_j}(x; \theta), y)],
\end{equation}
where $\mathcal{L}$ is the loss function, $x$ represents the features, $y$ represents the labels in $\mathcal{D}_T$, and $\theta$ denotes the parameters of $f_{T_j}$ that are being optimized. We calculate the expectation by the probability distribution $\mathcal{P}_T$, which has been closely aligned with $\mathcal{P}_k'$ to ensure consistency and maximize the efficacy of the knowledge transfer.

During update, the optimization of model parameters $\theta$ leverages both the target dataset $\mathcal{D}_T$ and the aligned source dataset $\mathcal{D}_k'$, incorporating domain-specific characteristics to reduce domain shift. We define the optimization process as:


\begin{align}
    \theta^{*} &= \arg\min_{\theta} \bigg(\alpha \mathbb{E}_{(x, y) \in \mathcal{D}_k'} [\mathcal{L}(f_{T_j}(x; \theta), y)] \nonumber \\
    &\quad + (1-\alpha) \mathbb{E}_{(x, y) \in \mathcal{D}_T} [\mathcal{L}(f_{T_j}(x; \theta), y)]\bigg).
\end{align}

This formula aims to fine-tune the decision function $f_{T_j}$ by minimizing the weighted sum of expected losses across the datasets. The loss function $\mathcal{L}(f_{T_j}(x; \theta), y)$ evaluates prediction accuracy, guiding the adjustment of parameters. The expectations $\mathbb{E}_{(x, y) \in \mathcal{D}_k'}$ and $\mathbb{E}_{(x, y) \in \mathcal{D}_T}$ represent the mean losses over the source and target datasets, respectively. The weighting factor $\alpha$ adjusts the relative influence of each dataset and enables flexible adaptation between leveraging established knowledge from the source and integrating new data from the target. In this way, we apply the transferred knowledge to enhance model performance on the target task.



Following the optimization of the model parameters, the backpropagation process then updates $\theta$ by minimizing the total loss. This loss is a weighted sum calculated from the losses on $\mathcal{D}_k'$ and $\mathcal{D}_T$. This involves calculating the gradient of the loss function with respect to $\theta$ and updating $\theta$ using gradient descent methods:
\begin{equation}
    \theta \leftarrow \theta - \eta \nabla_\theta \left(\alpha \mathcal{L}(\mathcal{D}_k'; \theta) + (1-\alpha) \mathcal{L}(\mathcal{D}_T; \theta)\right),
\end{equation}
where $\eta$ is the learning rate. The gradients are computed based on both datasets, which allows the model to learn from both the aligned source data and the target data.

After the parameter optimization and backpropagation, we evaluate the effectiveness of the transfer learning process; we adopt a performance metric $\phi$ on the target dataset:
\begin{equation}
    \phi(f_{T_j}^{*}, \mathcal{D}_T),
\end{equation}
where $\phi$ measures the performance of the optimized model $f_{T_j}^{*}$ on $\mathcal{D}_T$.

In this way, we systematically improve the model based on empirical performance metrics.

%% file: main/4-exp.tex


\begin{table}[t]
\centering
\fontsize{9}{11}\selectfont
\setlength{\tabcolsep}{8pt} 

\begin{tabular}{l|lll}
    \toprule
    Datasets & Samples & Features & Class \\
    \midrule
    BCW    & 570      & 30     & 2 \\
    VID    & 8631    & 22 & 50 \\
    HD    & 303    & 13      & 2 \\
    TCC    & 7043  & 21    & 2 \\
    \bottomrule
\end{tabular}
\caption{Datasets description. Here we use four datasets from the medical and economic domains.}
\label{tab:dataset}
\end{table}

\section{Experiments}
In this section, we present four experiments to demonstrate the effectiveness and impacts of the \model. First, we compare the \model against several baseline methods on four downstream tasks. Second, we present the correlations between several target domains and their retrieved source domains.  Finally, we discuss the reason for performance improvement.

\begin{table}[tbp] 
  \centering
  \label{tab:addlabel} 
  \resizebox{0.3\textwidth}{!}{ 
    {\fontsize{9pt}{11pt}\selectfont
    \begin{tabular}{|c|c|c|c|c|}
      \hline
      Dataset \rule{0pt}{10pt} & Metrics & FTT & TTab & LEKA \\
      \hline
      \multirow{4}[8]{*}[2ex]{BCW} & Acc \rule{0pt}{10pt}  & 0.956 & 0.956 & \textbf{0.991} \\
      \cline{2-5}          & Prec \rule{0pt}{10pt} & 0.951 & 0.948 & \textbf{0.988} \\
      \cline{2-5}          & Rec \rule{0pt}{10pt}  & 0.960 & 0.960 & \textbf{0.993} \\
      \cline{2-5}          & F1  \rule{0pt}{10pt}  & 0.955 & 0.954 & \textbf{0.990} \\
      \hline
      \multirow{4}[8]{*}[2ex]{VID} & Acc \rule{0pt}{10pt}  & 0.745 & 0.797 & \textbf{0.995} \\
      \cline{2-5}          & Prec \rule{0pt}{10pt} & 0.758 & 0.588 & \textbf{0.996} \\
      \cline{2-5}          & Rec \rule{0pt}{10pt}  & 0.747 & 0.513 & \textbf{0.996} \\
      \cline{2-5}          & F1  \rule{0pt}{10pt}  & 0.739 & 0.519 & \textbf{0.996} \\
      \hline
      \multirow{4}[8]{*}[2ex]{HD}  & Acc \rule{0pt}{10pt}  & 0.738 & 0.803 & \textbf{0.918} \\
      \cline{2-5}          & Prec \rule{0pt}{10pt} & 0.726 & 0.802 & \textbf{0.914} \\
      \cline{2-5}          & Rec \rule{0pt}{10pt}  & 0.718 & 0.802 & \textbf{0.918} \\
      \cline{2-5}          & F1 \rule{0pt}{10pt}   & 0.721 & 0.802 & \textbf{0.916} \\
      \hline
      \multirow{4}[8]{*}[2ex]{TCC} & Acc \rule{0pt}{10pt}  & 0.836 & 0.795 & \textbf{0.887} \\
      \cline{2-5}          & Prec \rule{0pt}{10pt} & 0.803 & 0.738 & \textbf{0.846} \\
      \cline{2-5}          & Rec  \rule{0pt}{10pt} & 0.865 & 0.712 & \textbf{0.901} \\
      \cline{2-5}          & F1  \rule{0pt}{10pt}  & 0.817 & 0.722 & \textbf{0.865} \\
      \hline
    \end{tabular}
    }
  }
  \caption{Performance comparison of transfer learning methods.}

\end{table}

\begin{figure}[tbp] 
  \centering
  \includegraphics[width=1\columnwidth]{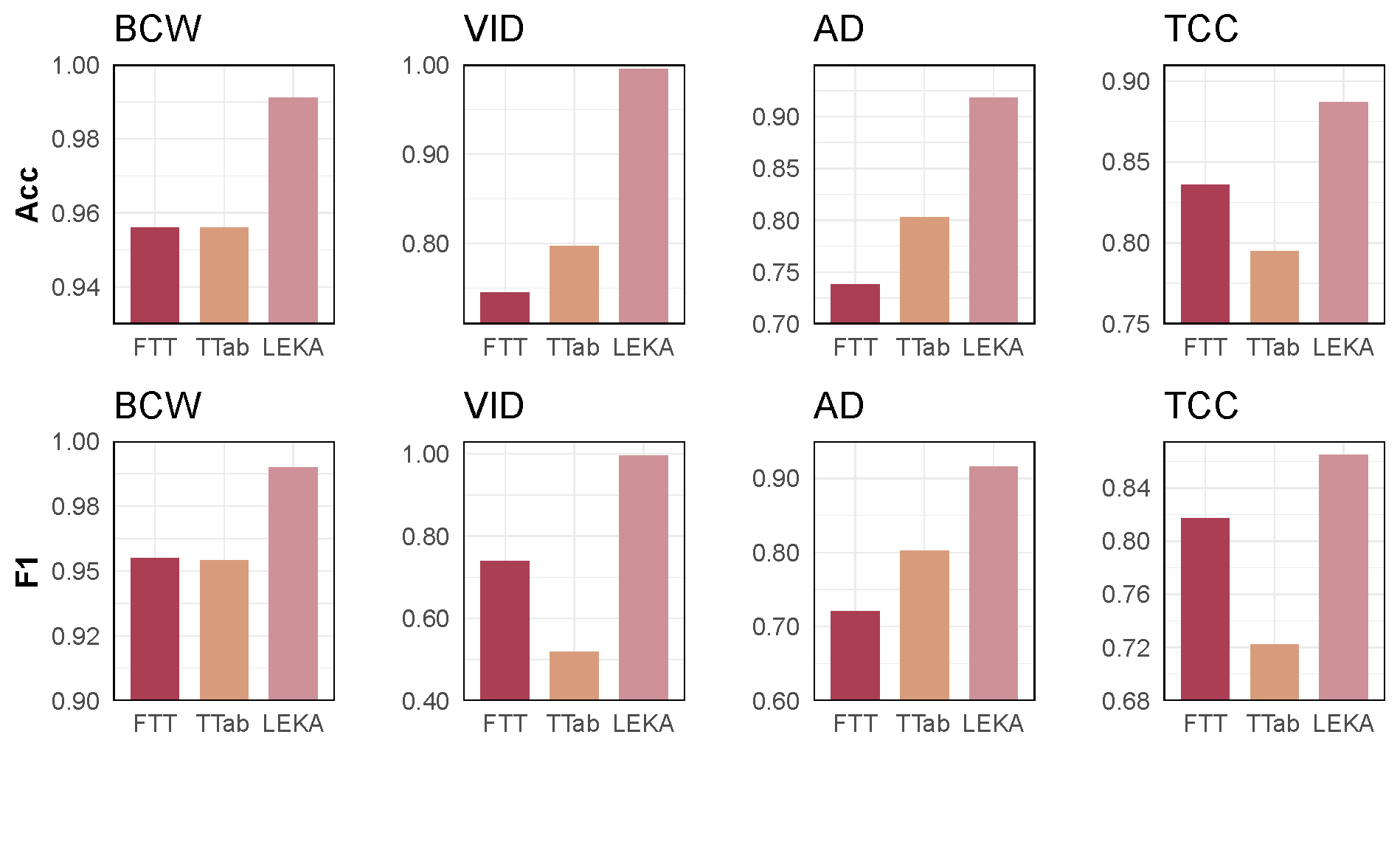} 
  \caption{Comparison of accuracy and F1 scores on various transfer learning methods.}
  \label{fig:accf1} 

\end{figure}

\subsection{Experiment Settings}
\paragraph{Datasets and Domains.} We evaluate our method on four datasets of medical and economic domains: (1) \textit{\textbf{Breast Cancer Wisconsin} (Diagnostic) (BCW)}~\cite{wolberg1995breast}, (2) \textit{\textbf{Heart Disease} (HD)}~\cite{janosi1988heart}, (3) \textit{\textbf{Vehicle Insurance Data} (VID)}~\cite{vehicle_insurance_data}, and (4) \textit{\textbf{Telco Customer Churn} (TCC)}~\cite{zhang2022data}. We show the detailed information about the features of the datasets in Table~\ref{tab:dataset}.

\paragraph{Metrics and Models.} We evaluate the model performance by the following metrics: \textit{Overall Accuracy (Acc)} measures the proportion of true results (both true positives and true negatives) in the total dataset. \textit{Precision (Prec)} reflects the ratio of true positive predictions to all positive predictions for each class. \textit{Recall (Rec)}, also known as sensitivity, reflects the ratio of true positive predictions to all actual positives for each class. \textit{F-Measure (F1)} is the harmonic mean of precision and recall, calculated here as the macro-average. We apply the \model across a range of models: 1) \textit{Tabnet (TN)}~\cite{arik2021tabnet}; 2) \textit{TabTransformer (TT)}~\cite{huang2020tabtransformer}; 3) \textit{Random Forest (RF)}~\cite{rigatti2017random}; 4) \textit{Gradient Boosting Decision Trees (GBDT)}~\cite{lin2023machine} ; 5) \textit{XGBoost (XB)}~\cite{chen2016xgboost}. We compare the performance in these tasks both with and without our method.

\paragraph{Baseline Models.} We compare the \model with six baseline methods, including: 1) \textbf{Raw Data}: using vanilla data; 2) \textbf{TIFG}~\cite{zhang2024tifg}: feature generation with LLM; 3) \textbf{KPDDS}~\cite{huang2024key}: data synthesis with LLM using key point examples; 4) \textbf{GReaT}~\cite{borisov2022language}: data synthesis with LLM  simulating data subset distributions; 5) \textbf{FTT}~\cite{levin2022transfer}: A TabTransformer pretrained on the source domain and then fine-tuned on the target domain; 6) \textbf{TTab}~\cite{wang2022transtab}: A transferable tabular Transformer capable of learning from multiple tabular datasets.

\paragraph{Implementation Details.} In our setup for data synthesis and model training, we utilize \texttt{GPT-4o}~\cite{openai2024gpt4o} as the query generator, combined with the \texttt{Exa API}~\cite{exa2024api} to fetch web pages containing datasets from \texttt{Kaggle}~\cite{kaggle2024} and the \texttt{UCI Machine Learning Repository}~\cite{uci2024repository} that may be suitable for knowledge transfer. We extract dataset descriptions from web pages, and \texttt{GPT-4o} assesses their potential for knowledge transfer. Additionally, \texttt{GPT-4o} serves as a generator for executable code, performing up to five code generations. For our models, we configure TN, TT, and FTT with a batch size of $512$ for the VID and TCC datasets and a batch size of $32$ for the BCW dataset, a maximum of $100$ epochs, and employ early stopping with a patience of $20$. The learning rate is set at the default $0.02$ for \texttt{pytorch\_tabnet}. For the RF and GBDT models, the number of trees is set to $100$, with GBDT also configured with a learning rate of $0.1$ and a max depth of $3$. TTab is set with a maximum of $50$ epochs, a learning rate of $1 \times 10^{-3}$, and a weight decay of $1 \times 10^{-4}$.

\begin{table*}
  \centering

  \resizebox{1\textwidth}{!}{
        \begin{tabular}{|c|l|rrrr|rrrr|rrrr|rrrr|rrrr|}
    \hline
    \multirow{3}[2]{*}{\textbf{Metrics}} & \multicolumn{1}{c|}{\multirow{3}[2]{*}{\textbf{Model}}} & \multicolumn{4}{c|}{\multirow{3}[2]{*}{\textbf{GBDT}}} & \multicolumn{4}{c|}{\multirow{3}[2]{*}{\textbf{RF}}} & \multicolumn{4}{c|}{\multirow{3}[2]{*}{\textbf{TN}}} & \multicolumn{4}{c|}{\multirow{3}[2]{*}{\textbf{XB}}} & \multicolumn{4}{c|}{\multirow{3}[2]{*}{\textbf{TT}}} \\
          &       & \multicolumn{4}{c|}{}         & \multicolumn{4}{c|}{}         & \multicolumn{4}{c|}{}         & \multicolumn{4}{c|}{}         & \multicolumn{4}{c|}{} \\
          &       & \multicolumn{4}{c|}{}         & \multicolumn{4}{c|}{}         & \multicolumn{4}{c|}{}         & \multicolumn{4}{c|}{}         & \multicolumn{4}{c|}{} \\
    \hline
    \multirow{6}[11]{*}[2ex]{\textbf{Acc}} & \rule{0pt}{10pt} & \multicolumn{1}{c}{BCW} & \multicolumn{1}{c}{VID} & \multicolumn{1}{c}{HD} & \multicolumn{1}{c|}{TCC} & \multicolumn{1}{c}{BCW} & \multicolumn{1}{c}{VID} & \multicolumn{1}{c}{HD} & \multicolumn{1}{c|}{TCC} & \multicolumn{1}{c}{BCW} & \multicolumn{1}{c}{VID} & \multicolumn{1}{c}{HD} & \multicolumn{1}{c|}{TCC} & \multicolumn{1}{c}{BCW} & \multicolumn{1}{c}{VID} & \multicolumn{1}{c}{HD} & \multicolumn{1}{c|}{TCC} & \multicolumn{1}{c}{BCW} & \multicolumn{1}{c}{VID} & \multicolumn{1}{c}{HD} & \multicolumn{1}{c|}{TCC} \\
\cline{2-22}          & Raw \rule{0pt}{10pt}  & 0.939  & 0.991  & 0.754  & 0.793  & 0.921  & 0.827  & 0.770  & 0.776  & 0.956  & 0.436  & 0.803  & 0.783  & 0.930  & 0.946  & 0.770  & 0.773  & 0.965  & 0.714  & 0.574  & 0.754  \\
\cline{2-2}          & KPDDS \rule{0pt}{10pt} & 0.947  & 0.992  & 0.787  & 0.787  & 0.939  & 0.846  & 0.803  & 0.787  & 0.982  & 0.567  & 0.820  & 0.806  & 0.956  & 0.948  & 0.836  & 0.789  & 0.974  & 0.728  & 0.639  & 0.781  \\
\cline{2-2}          & GReaT \rule{0pt}{10pt} & \underline{0.974}  & 0.994  & 0.803  & 0.808  & 0.947  & 0.847  & 0.852  & 0.788  & 0.965  & 0.506  & 0.885  & 0.795  & \underline{0.974}  & \underline{0.953}  & 0.820  & 0.784  & \underline{0.974}  & 0.724  & 0.656  & 0.754  \\
\cline{2-2}          & TIFG \rule{0pt}{10pt} & 0.947  & \underline{0.994}  & \underline{0.836}  & \underline{0.815}  & \underline{0.947}  & \underline{0.858}  & \underline{0.869}  & \underline{0.800}  & \underline{0.982}  & \underline{0.579}  & \underline{0.902}  & \underline{0.815}  & 0.965  & 0.952  & \underline{0.852}  & \underline{0.803}  & 0.965  & \underline{0.736}  & \underline{0.705}  & \underline{0.790}  \\
\cline{2-2}          & LEKA \rule{0pt}{10pt} & \textbf{0.991} & \textbf{0.995} & \textbf{0.869} & \textbf{0.825} & \textbf{0.982} & \textbf{0.868} & \textbf{0.885} & \textbf{0.811} & \textbf{0.982} & \textbf{0.596} & \textbf{0.918} & \textbf{0.838} & \textbf{0.991} & \textbf{0.960} & \textbf{0.885} & \textbf{0.823} & \textbf{0.991} & \textbf{0.787} & \textbf{0.754} & \textbf{0.887} \\
\hline
    \multirow{6}[10]{*}[2ex]{\textbf{Prec}} & \rule{0pt}{10pt} & \multicolumn{1}{c}{BCW} & \multicolumn{1}{c}{VID} & \multicolumn{1}{c}{HD} & \multicolumn{1}{c|}{TCC} & \multicolumn{1}{c}{BCW} & \multicolumn{1}{c}{VID} & \multicolumn{1}{c}{HD} & \multicolumn{1}{c|}{TCC} & \multicolumn{1}{c}{BCW} & \multicolumn{1}{c}{VID} & \multicolumn{1}{c}{HD} & \multicolumn{1}{c|}{TCC} & \multicolumn{1}{c}{BCW} & \multicolumn{1}{c}{VID} & \multicolumn{1}{c}{HD} & \multicolumn{1}{c|}{TCC} & \multicolumn{1}{c}{BCW} & \multicolumn{1}{c}{VID} & \multicolumn{1}{c}{HD} & \multicolumn{1}{c|}{TCC} \\
\cline{2-22}          & Raw \rule{0pt}{10pt}  & 0.948  & 0.992  & 0.804  & 0.734  & 0.926  & 0.834  & 0.751  & 0.716  & 0.949  & 0.429  & 0.803  & 0.717  & 0.928  & 0.932  & 0.775  & 0.706  & 0.965  & 0.742  & 0.302  & 0.709  \\
\cline{2-2}          & KPDDS \rule{0pt}{10pt} & 0.936  & 0.993  & 0.782  & 0.720  & 0.929  & 0.850  & 0.807  & 0.725  & 0.978  & 0.594  & 0.823  & \underline{0.767}  & 0.954  & 0.934  & 0.843  & 0.744  & 0.974  & \underline{0.752}  & \textbf{0.817} & 0.724  \\
\cline{2-2}          & GReaT \rule{0pt}{10pt} & \underline{0.974}  & 0.994  & 0.804  & 0.752  & \underline{0.949}  & 0.851  & 0.852  & 0.743  & 0.959  & 0.448  & 0.895  & 0.746  & 0.966  & 0.955  & 0.823  & \underline{0.748}  & \underline{0.979}  & 0.732  & 0.659  & 0.697  \\
\cline{2-2}          & TIFG \rule{0pt}{10pt}  & 0.947  & \underline{0.994}  & \underline{0.840}  & \underline{0.767}  & 0.940  & \underline{0.863}  & \underline{0.867}  & \underline{0.750}  & \underline{0.981}  & \underline{0.571}  & \underline{0.900}  & 0.748  & \underline{0.967}  & \underline{0.957}  & \underline{0.849}  & 0.750  & 0.965  & 0.725  & 0.707  & \underline{0.734}  \\
\cline{2-2}          & LEKA \rule{0pt}{10pt} & \textbf{0.988} & \textbf{0.996} & \textbf{0.863} & \textbf{0.782} & \textbf{0.980} & \textbf{0.869} & \textbf{0.887} & \textbf{0.760} & \textbf{0.977} & \textbf{0.619} & \textbf{0.914} & \textbf{0.823} & \textbf{0.989} & \textbf{0.962} & \textbf{0.885} & \textbf{0.785} & \textbf{0.991} & \textbf{0.802} & \underline{0.752}  & \textbf{0.846} \\
\hline
    \multirow{6}[10]{*}[2ex]{\textbf{Rec}} & \rule{0pt}{10pt} & \multicolumn{1}{c}{BCW} & \multicolumn{1}{c}{VID} & \multicolumn{1}{c}{HD} & \multicolumn{1}{c|}{TCC} & \multicolumn{1}{c}{BCW} & \multicolumn{1}{c}{VID} & \multicolumn{1}{c}{HD} & \multicolumn{1}{c|}{TCC} & \multicolumn{1}{c}{BCW} & \multicolumn{1}{c}{VID} & \multicolumn{1}{c}{HD} & \multicolumn{1}{c|}{TCC} & \multicolumn{1}{c}{BCW} & \multicolumn{1}{c}{VID} & \multicolumn{1}{c}{HD} & \multicolumn{1}{c|}{TCC} & \multicolumn{1}{c}{BCW} & \multicolumn{1}{c}{VID} & \multicolumn{1}{c}{HD} & \multicolumn{1}{c|}{TCC} \\
\cline{2-22}          & Raw \rule{0pt}{10pt}  & 0.912  & 0.991  & 0.770  & 0.701  & 0.918  & 0.828  & 0.751  & 0.671  & 0.954  & 0.443  & 0.803  & 0.680  & 0.928  & 0.926  & 0.769  & 0.682  & 0.965  & 0.720  & 0.461  & 0.757  \\
\cline{2-2}          & KPDDS \rule{0pt}{10pt} & 0.959  & 0.993  & 0.785  & 0.689  & \underline{0.951}  & 0.845  & 0.806  & 0.675  & \underline{0.986}  & 0.553  & 0.819  & 0.686  & 0.962  & 0.930  & 0.845  & 0.710  & 0.974  & 0.734  & 0.522  & 0.755  \\
\cline{2-2}          & GReaT \rule{0pt}{10pt} & \underline{0.968}  & 0.994  & 0.804  & 0.717  & 0.944  & 0.846  & 0.853  & \underline{0.702}  & 0.967  & 0.502  & 0.887  & 0.692  & \underline{0.973}  & \underline{0.954}  & 0.827  & 0.704  & \underline{0.968}  & \underline{0.726}  & 0.661  & 0.717  \\
\cline{2-2}          & TIFG \rule{0pt}{10pt} & 0.935  & \underline{0.994}  & \underline{0.839}  & \underline{0.725}  & 0.940  & \underline{0.856}  & \underline{0.867}  & 0.697  & 0.981  & \underline{0.570}  & \underline{0.904}  & \underline{0.700}  & 0.957  & 0.951  & \underline{0.852}  & \underline{0.711}  & 0.965  & 0.733  & \underline{0.690}  & \underline{0.724}  \\
\cline{2-2}          & LEKA \rule{0pt}{10pt} & \textbf{0.993} & \textbf{0.996} & \textbf{0.877} & \textbf{0.740} & \textbf{0.985} & \textbf{0.865} & \textbf{0.887} & \textbf{0.725} & \textbf{0.986} & \textbf{0.603} & \textbf{0.918} & \textbf{0.732} & \textbf{0.993} & \textbf{0.958} & \textbf{0.885} & \textbf{0.747} & \textbf{0.991} & \textbf{0.789} & \textbf{0.768} & \textbf{0.901} \\
\hline
    \multirow{6}[11]{*}[2ex]{\textbf{F1}} & \rule{0pt}{10pt} & \multicolumn{1}{c}{BCW} & \multicolumn{1}{c}{VID} & \multicolumn{1}{c}{HD} & \multicolumn{1}{c|}{TCC} & \multicolumn{1}{c}{BCW} & \multicolumn{1}{c}{VID} & \multicolumn{1}{c}{HD} & \multicolumn{1}{c|}{TCC} & \multicolumn{1}{c}{BCW} & \multicolumn{1}{c}{VID} & \multicolumn{1}{c}{HD} & \multicolumn{1}{c|}{TCC} & \multicolumn{1}{c}{BCW} & \multicolumn{1}{c}{VID} & \multicolumn{1}{c}{HD} & \multicolumn{1}{c|}{TCC} & \multicolumn{1}{c}{BCW} & \multicolumn{1}{c}{VID} & \multicolumn{1}{c}{HD} & \multicolumn{1}{c|}{TCC} \\
\cline{2-22}          & Raw \rule{0pt}{10pt}  & 0.927  & 0.991  & 0.750  & 0.714  & 0.920  & 0.827  & 0.751  & 0.685  & 0.952  & 0.400  & 0.803  & 0.693  & 0.928  & 0.927  & 0.769  & 0.692  & 0.965  & 0.704  & 0.365  & 0.718  \\
\cline{2-2}          & KPDDS \rule{0pt}{10pt} & 0.944  & 0.993  & 0.783  & 0.701  & 0.936  & 0.844  & 0.803  & 0.691  & 0.982  & 0.510  & 0.819  & 0.708  & 0.956  & 0.930  & 0.836  & 0.723  & \underline{0.974}  & \underline{0.724}  & 0.429  & \underline{0.735}  \\
\cline{2-2}          & GReaT \rule{0pt}{10pt} & 0.971  & 0.994  & 0.803  & 0.731  & \underline{0.946}  & 0.846  & 0.852  & \underline{0.716}  & 0.963  & 0.440  & 0.885  & 0.709  & \underline{0.970}  & \underline{0.952}  & 0.819  & 0.718  & 0.973  & 0.712  & 0.655  & 0.704  \\
\cline{2-2}          & TIFG \rule{0pt}{10pt} & \underline{0.941}  & \underline{0.994}  & \underline{0.836}  & \underline{0.741}  & 0.940  & \underline{0.853}  & \underline{0.867}  & 0.715  & \underline{0.981}  & \underline{0.535}  & \underline{0.901}  & \underline{0.718}  & 0.961  & 0.951  & \underline{0.850}  & \underline{0.726}  & 0.965  & 0.706  & \underline{0.691}  & 0.729  \\
\cline{2-2}          & LEKA \rule{0pt}{10pt} & \textbf{0.990} & \textbf{0.996} & \textbf{0.866} & \textbf{0.756} & \textbf{0.982} & \textbf{0.864} & \textbf{0.885} & \textbf{0.739} & \textbf{0.981} & \textbf{0.568} & \textbf{0.916} & \textbf{0.760} & \textbf{0.991} & \textbf{0.959} & \textbf{0.885} & \textbf{0.761} & \textbf{0.991} & \textbf{0.780} & \textbf{0.750} & \textbf{0.865} \\
    \hline
    \end{tabular}%
    }

\caption{Overall performance on downstream tasks. The best results are highlighted in \textbf{bold}, and the runner-up results are highlighted in \underline{underline}. (Higher values indicate better performance.)}
\label{tab:overall}
\end{table*}%

\subsection{Experiment Results}
\paragraph{Overall Performance.} 
The results are shown in Table~\ref{tab:overall}, comparing our \model across four datasets. In summary:

(1) Compared with baseline models, \model significantly improves accuracy, surpassing baseline methods like TIFG and GReaT, with enhancements up to $4.4\%$ in models like GBDT. \model consistently achieves top results in precision and recall, indicating its precision in correctly identifying relevant cases while minimizing false positives. The F1 scores under \model are notably high, reflecting its effective balance between precision and recall across various models and datasets. These results validate the effectiveness of \model’s retrieval and harmonization strategies and its robustness in diverse application scenarios. Overall, \model's strategic approach to knowledge transfer is particularly advantageous in complex data environments, showcasing its adaptability and efficiency compared to traditional methods.

(2) The \model outperforms all baseline methods in almost all metrics and datasets. Specifically, \model shows enhancements in accuracy by $2-5\%$ over other methods. It demonstrates notable improvements of up to $4.4\%$ in GBDT for the VID dataset and $5.4\%$ in RF for the BCW dataset. For improvement of overall accuracy, we demonstrate the \model's effective retrieval and harmonization of feature space and marginal probability measures with the target datasets. 
\model demonstrates exceptional precision, improving the TT metric for the TCC dataset by over $10\%$ compared to baselines, effectively reducing false positives.
Meanwhile, the \model consistently outperforms baselines in reducing misclassification rates with the highest F1 scores. These results prove that \model's retrieval and data harmonization reduce misclassification by forming a deeper and clearer understanding of the potential relationship between features.


\paragraph{Comparison with transfer learning methods.} We then compare our \model method with transfer learning methods to demonstrate its effectiveness in enhancing model performance in complex domains. In these scenarios, the transfer learning methods show their reliance on manual data selection and alignment processes. The results demonstrate that \model outperforms transfer learning methods, FFT, and TTab across various metrics and datasets. For example, in the BCW dataset, LEKA improves accuracy by $3.5\%$ and recall by $3.3\%$ compared to the competing methods. This superior performance is consistent across other datasets like VID, HD, and TCC, with notable enhancements in precision and recall, emphasizing LEKA's effective data harmonization capabilities. This adaptability and efficiency in handling diverse datasets underscore LEKA's robustness and advantages.

%% file: main/5-concl.tex
\section{Conclusion}

In this paper, we introduce LLM-Enhanced Knowledge Augmentation (LEKA), a novel retrieval and harmonization framework that dynamically refines source data for effective knowledge transfer. This structure significantly enhances data augmentation by leveraging advanced LLM capabilities to automatically align and optimize data retrieved from diverse external libraries. Extensive experiments across various tasks demonstrate superiority to existing methods, especially in improving model adaptability and accuracy in complex data environments. By adopting LEKA on data-scarce domains, we achieve substantial improvements in learning performance and domain-specific task accuracy. For future work, we plan to extend this framework to include more varied data types and conduct a thorough empirical analysis to understand its underlying mechanisms and impacts.

\section{Limitations}
We acknowledge the following limitations: (1) the current work has only been tested on tabular data, and more complex test scenarios have not yet been involved; (2) despite its automation, \model can be computationally intensive. This could limit its applicability in resource-constrained environments or require substantial computational resources to maintain operational efficiency; (3) adopting the \model in tasks with unique requirements may have inherent limitations.


%% file: main/7-acknowledgment.tex
\section*{Acknowledgements}
Dr. Kunpeng Liu is supported by the National Science Foundation (NSF) via the grant numbers 2426339 and 2348485. Dr. Yanjie Fu is supported by the National Science Foundation (NSF) via the grant numbers 2426340, 2416727, 2421864, 2421865, 2421803, and National Academy of Engineering Grainger Foundation Frontiers of Engineering Grants.

%% file: ijcai25.bib
@article{zhang2024tifg,
  title={TIFG: Text-Informed Feature Generation with Large Language Models},
  author={Zhang, Xinhao and Zhang, Jinghan and Mo, Fengran and Chen, Yuzhong and Liu, Kunpeng},
  journal={arXiv preprint arXiv:2406.11177},
  year={2024}
}

@article{guo2017deepfm,
  title={DeepFM: a factorization-machine based neural network for CTR prediction},
  author={Guo, Huifeng and Tang, Ruiming and Ye, Yunming and Li, Zhenguo and He, Xiuqiang},
  journal={arXiv preprint arXiv:1703.04247},
  year={2017}
}

@article{lewis2020retrieval,
  title={Retrieval-augmented generation for knowledge-intensive nlp tasks},
  author={Lewis, Patrick and Perez, Ethan and Piktus, Aleksandra and Petroni, Fabio and Karpukhin, Vladimir and Goyal, Naman and K{\"u}ttler, Heinrich and Lewis, Mike and Yih, Wen-tau and Rockt{\"a}schel, Tim and others},
  journal={Advances in Neural Information Processing Systems},
  volume={33},
  pages={9459--9474},
  year={2020}
}

@article{gao2023retrieval,
  title={Retrieval-augmented generation for large language models: A survey},
  author={Gao, Yunfan and Xiong, Yun and Gao, Xinyu and Jia, Kangxiang and Pan, Jinliu and Bi, Yuxi and Dai, Yi and Sun, Jiawei and Wang, Haofen},
  journal={arXiv preprint arXiv:2312.10997},
  year={2023}
}

@article{zhang2024dynamic,
  title={Dynamic and Adaptive Feature Generation with LLM},
  author={Zhang, Xinhao and Zhang, Jinghan and Rekabdar, Banafsheh and Zhou, Yuanchun and Wang, Pengfei and Liu, Kunpeng},
  journal={arXiv preprint arXiv:2406.03505},
  year={2024}
}

@article{hu2024rag,
  title={Rag and rau: A survey on retrieval-augmented language model in natural language processing},
  author={Hu, Yucheng and Lu, Yuxing},
  journal={arXiv preprint arXiv:2404.19543},
  year={2024}
}

@article{liu2021automated,
  title={Automated feature selection: A reinforcement learning perspective},
  author={Liu, Kunpeng and Fu, Yanjie and Wu, Le and Li, Xiaolin and Aggarwal, Charu and Xiong, Hui},
  journal={IEEE Transactions on Knowledge and Data Engineering},
  volume={35},
  number={3},
  pages={2272--2284},
  year={2021},
  publisher={IEEE}
}

@article{cao2009agent,
  title={Agent mining: The synergy of agents and data mining},
  author={Cao, Longbing and Gorodetsky, Vladimir and Mitkas, Pericles A},
  journal={IEEE intelligent systems},
  volume={24},
  number={3},
  pages={64--72},
  year={2009},
  publisher={IEEE}
}

@article{durante2024agent,
  title={Agent ai: Surveying the horizons of multimodal interaction},
  author={Durante, Zane and Huang, Qiuyuan and Wake, Naoki and Gong, Ran and Park, Jae Sung and Sarkar, Bidipta and Taori, Rohan and Noda, Yusuke and Terzopoulos, Demetri and Choi, Yejin and others},
  journal={arXiv preprint arXiv:2401.03568},
  year={2024}
}

@article{alyafeai2020survey,
  title={A survey on transfer learning in natural language processing},
  author={Alyafeai, Zaid and AlShaibani, Maged Saeed and Ahmad, Irfan},
  journal={arXiv preprint arXiv:2007.04239},
  year={2020}
}

@article{han2021robust,
  title={Robust transfer learning with pretrained language models through adapters},
  author={Han, Wenjuan and Pang, Bo and Wu, Yingnian},
  journal={arXiv preprint arXiv:2108.02340},
  year={2021}
}

@article{yordanov2021few,
  title={Few-shot out-of-domain transfer learning of natural language explanations in a label-abundant setup},
  author={Yordanov, Yordan and Kocijan, Vid and Lukasiewicz, Thomas and Camburu, Oana-Maria},
  journal={arXiv preprint arXiv:2112.06204},
  year={2021}
}

@article{siriwardhana2023improving,
  title={Improving the domain adaptation of retrieval augmented generation (RAG) models for open domain question answering},
  author={Siriwardhana, Shamane and Weerasekera, Rivindu and Wen, Elliott and Kaluarachchi, Tharindu and Rana, Rajib and Nanayakkara, Suranga},
  journal={Transactions of the Association for Computational Linguistics},
  volume={11},
  pages={1--17},
  year={2023},
  publisher={MIT Press One Broadway, 12th Floor, Cambridge, Massachusetts 02142, USA~…}
}

@article{fleischer2024rag,
  title={RAG Foundry: A Framework for Enhancing LLMs for Retrieval Augmented Generation},
  author={Fleischer, Daniel and Berchansky, Moshe and Wasserblat, Moshe and Izsak, Peter},
  journal={arXiv preprint arXiv:2408.02545},
  year={2024}
}

@article{feng2021survey,
  title={A survey of data augmentation approaches for NLP},
  author={Feng, Steven Y and Gangal, Varun and Wei, Jason and Chandar, Sarath and Vosoughi, Soroush and Mitamura, Teruko and Hovy, Eduard},
  journal={arXiv preprint arXiv:2105.03075},
  year={2021}
}

@inproceedings{arik2021tabnet,
  title={Tabnet: Attentive interpretable tabular learning},
  author={Arik, Sercan {\"O} and Pfister, Tomas},
  booktitle={Proceedings of the AAAI conference on artificial intelligence},
  volume={35(8)},
  pages={6679--6687},
  year={2021}
}

@article{rigatti2017random,
  title={Random forest},
  author={Rigatti, Steven J},
  journal={Journal of Insurance Medicine},
  volume={47},
  number={1},
  pages={31--39},
  year={2017},
  publisher={American Academy of Insurance Medicine 1700 Magnavox Way, Fort Wayne, IN 46804}
}

@inproceedings{lin2023machine,
  title={Machine unlearning in gradient boosting decision trees},
  author={Lin, Huawei and Chung, Jun Woo and Lao, Yingjie and Zhao, Weijie},
  booktitle={Proceedings of the 29th ACM SIGKDD Conference on Knowledge Discovery and Data Mining},
  pages={1374--1383},
  year={2023}
}

@article{huang2020tabtransformer,
  title={Tabtransformer: Tabular data modeling using contextual embeddings},
  author={Huang, Xin and Khetan, Ashish and Cvitkovic, Milan and Karnin, Zohar},
  journal={arXiv preprint arXiv:2012.06678},
  year={2020}
}

@misc{wolberg1995breast,
  author       = {Wolberg, William and Mangasarian, Olvi and Street, Nick and Street, W.},
  title        = {{Breast Cancer Wisconsin (Diagnostic)}},
  year         = {1995},
  howpublished = {\url{https://archive.ics.uci.edu/dataset/17/breast+cancer+wisconsin+diagnostic}},
  note         = {Accessed: 2025-01-21}
}

@misc{vehicle_insurance_data,
  author       = {Himanshu Bhatt},
  title        = {{Vehicle Insurance Data}},
  year         = {2019},
  howpublished = {\url{https://www.kaggle.com/datasets/junglisher/vehicle-insurance-data}},
  note         = {Accessed: 2025-01-21}
}

@misc{janosi1988heart,
  author       = {Janosi, Andras and Steinbrunn, William and Pfisterer, Matthias and Detrano, Robert},
  title        = {{Heart Disease}},
  year         = {1989},
  howpublished = {\url{https://archive.ics.uci.edu/dataset/45/heart+disease}},
  note         = {Accessed: 2025-01-21}
}

@misc{zhang2022data,
  author       = {BlastChar},
  title        = {{Telco Customer Churn}},
  year         = {2018},
  howpublished = {\url{https://www.kaggle.com/datasets/blastchar/telco-customer-churn}},
  note         = {Accessed: 2025-01-21}
}

@inproceedings{chen2016xgboost,
  title={Xgboost: A scalable tree boosting system},
  author={Chen, Tianqi and Guestrin, Carlos},
  booktitle={Proceedings of the 22nd acm sigkdd international conference on knowledge discovery and data mining},
  pages={785--794},
  year={2016}
}

@article{huang2024key,
  title={Key-point-driven data synthesis with its enhancement on mathematical reasoning},
  author={Huang, Yiming and Liu, Xiao and Gong, Yeyun and Gou, Zhibin and Shen, Yelong and Duan, Nan and Chen, Weizhu},
  journal={arXiv preprint arXiv:2403.02333},
  year={2024}
}

@article{borisov2022language,
  title={Language models are realistic tabular data generators},
  author={Borisov, Vadim and Se{\ss}ler, Kathrin and Leemann, Tobias and Pawelczyk, Martin and Kasneci, Gjergji},
  journal={arXiv preprint arXiv:2210.06280},
  year={2022}
}

@article{levin2022transfer,
  title={Transfer learning with deep tabular models},
  author={Levin, Roman and Cherepanova, Valeriia and Schwarzschild, Avi and Bansal, Arpit and Bruss, C Bayan and Goldstein, Tom and Wilson, Andrew Gordon and Goldblum, Micah},
  journal={arXiv preprint arXiv:2206.15306},
  year={2022}
}

@article{wang2022transtab,
  title={Transtab: Learning transferable tabular transformers across tables},
  author={Wang, Zifeng and Sun, Jimeng},
  journal={Advances in Neural Information Processing Systems},
  volume={35},
  pages={2902--2915},
  year={2022}
}

@inproceedings{tang2020onlineaugment,
  title={OnlineAugment: Online data augmentation with less domain knowledge},
  author={Tang, Zhiqiang and Gao, Yunhe and Karlinsky, Leonid and Sattigeri, Prasanna and Feris, Rogerio and Metaxas, Dimitris},
  booktitle={Computer Vision--ECCV 2020: 16th European Conference, Glasgow, UK, August 23--28, 2020, Proceedings, Part VII 16},
  pages={313--329},
  year={2020},
  organization={Springer}
}

@article{SHI201881,
title = {An efficient feature generation approach based on deep learning and feature selection techniques for traffic classification},
journal = {Computer Networks},
volume = {132},
pages = {81-98},
year = {2018},
issn = {1389-1286},
doi = {https://doi.org/10.1016/j.comnet.2018.01.007},
url = {https://www.sciencedirect.com/science/article/pii/S1389128618300082},
author = {Hongtao Shi and Hongping Li and Dan Zhang and Chaqiu Cheng and Xuanxuan Cao}
}

@article{zhang2024raft,
  title={Raft: Adapting language model to domain specific rag},
  author={Zhang, Tianjun and Patil, Shishir G and Jain, Naman and Shen, Sheng and Zaharia, Matei and Stoica, Ion and Gonzalez, Joseph E},
  journal={arXiv preprint arXiv:2403.10131},
  year={2024}
}

@article{huang2024survey,
  title={A Survey on Retrieval-Augmented Text Generation for Large Language Models},
  author={Huang, Yizheng and Huang, Jimmy},
  journal={arXiv preprint arXiv:2404.10981},
  year={2024}
}

@inproceedings{ringwald2021adaptiope,
  title={Adaptiope: A modern benchmark for unsupervised domain adaptation},
  author={Ringwald, Tobias and Stiefelhagen, Rainer},
  booktitle={Proceedings of the IEEE/CVF winter conference on applications of computer vision},
  pages={101--110},
  year={2021}
}

@article{nam2024tabular,
  title={Tabular Transfer Learning via Prompting LLMs},
  author={Nam, Jaehyun and Song, Woomin and Park, Seong Hyeon and Tack, Jihoon and Yun, Sukmin and Kim, Jaehyung and Oh, Kyu Hwan and Shin, Jinwoo},
  journal={arXiv preprint arXiv:2408.11063},
  year={2024}
}

@article{seemakhupt2024edgerag,
  title={EdgeRAG: Online-Indexed RAG for Edge Devices},
  author={Seemakhupt, Korakit and Liu, Sihang and Khan, Samira},
  journal={arXiv preprint arXiv:2412.21023},
  year={2024}
}

@article{jin2024ragcache,
  title={RAGCache: Efficient Knowledge Caching for Retrieval-Augmented Generation},
  author={Jin, Chao and Zhang, Zili and Jiang, Xuanlin and Liu, Fangyue and Liu, Xin and Liu, Xuanzhe and Jin, Xin},
  journal={arXiv preprint arXiv:2404.12457},
  year={2024}
}

@inproceedings{zhang2025ratt,
  title={RATT: A Thought Structure for Coherent and Correct LLM Reasoning},
  author={Zhang, Jinghan and Wang, Xiting and Ren, Weijieying and Jiang, Lu and Wang, Dongjie and Liu, Kunpeng},
  booktitle={Proceedings of the AAAI Conference on Artificial Intelligence},
  volume={39(25)},
  pages={26733--26741},
  year={2025}
}

@article{zhang2024prototypical,
  title={Prototypical reward network for data-efficient rlhf},
  author={Zhang, Jinghan and Wang, Xiting and Jin, Yiqiao and Chen, Changyu and Zhang, Xinhao and Liu, Kunpeng},
  journal={arXiv preprint arXiv:2406.06606},
  year={2024}
}

@misc{zhang2025blindspotnavigationllm,
      title={Blind Spot Navigation in LLM Reasoning with Thought Space Explorer}, 
      author={Jinghan Zhang and Fengran Mo and Xiting Wang and Kunpeng Liu},
      year={2025},
      eprint={2410.24155},
      archivePrefix={arXiv},
      primaryClass={cs.CL},
      url={https://arxiv.org/abs/2410.24155}, 
}

@article{wang2025diversity,
  title={Diversity-Oriented Data Augmentation with Large Language Models},
  author={Wang, Zaitian and Zhang, Jinghan and Zhang, Xinhao and Liu, Kunpeng and Wang, Pengfei and Zhou, Yuanchun},
  journal={arXiv preprint arXiv:2502.11671},
  year={2025}
}

@article{xie2024scoring,
  title={Scoring with Large Language Models: A Study on Measuring Empathy of Responses in Dialogues},
  author={Xie, Henry J and Zhang, Jinghan and Zhang, Xinhao and Liu, Kunpeng},
  journal={arXiv preprint arXiv:2412.20264},
  year={2024}
}

@article{wang2022semi,
  title={Semi-supervised learning for k-dependence Bayesian classifiers},
  author={Wang, Limin and Zhang, Xinhao and Li, Kuo and Zhang, Shuai},
  journal={Applied Intelligence},
  pages={1--19},
  year={2022},
  publisher={Springer}
}

@article{khodaee2024knowledge,
  title={Knowledge transfer in lifelong machine learning: a systematic literature review},
  author={Khodaee, Pouya and Viktor, Herna L and Michalowski, Wojtek},
  journal={Artificial Intelligence Review},
  volume={57},
  number={8},
  pages={217},
  year={2024},
  publisher={Springer}
}

@article{edge2024local,
  title={From local to global: A graph rag approach to query-focused summarization},
  author={Edge, Darren and Trinh, Ha and Cheng, Newman and Bradley, Joshua and Chao, Alex and Mody, Apurva and Truitt, Steven and Metropolitansky, Dasha and Ness, Robert Osazuwa and Larson, Jonathan},
  journal={arXiv preprint arXiv:2404.16130},
  year={2024}
}

@article{li2025oreo,
  title={Oreo: A Plug-in Context Reconstructor to Enhance Retrieval-Augmented Generation},
  author={Li, Sha and Ramakrishnan, Naren},
  journal={arXiv preprint arXiv:2502.13019},
  year={2025}
}

@misc{openai2024gpt4o,
  author={{OpenAI}},
  year={2024},
  title={{GPT-4o}},
  howpublished={\url{https://platform.openai.com/docs/models/gpt-4o}},
  note={Accessed: 2025-01-10}
}

@misc{exa2024api,
  author={{Exa}},
  year={2024},
  title={{Exa API}},
  howpublished={\url{https://exa.ai/}},
  note={Accessed: 2025-01-10}
}

@misc{kaggle2024,
  author={{Kaggle}},
  year={2024},
  title={{Kaggle: Your Machine Learning and Data Science Community}},
  howpublished={\url{https://www.kaggle.com/}},
  note={Accessed: 2025-01-10}
}

@misc{uci2024repository,
  author={{University of California, Irvine}},
  year={2024},
  title={{UCI Machine Learning Repository}},
  howpublished={\url{https://archive.ics.uci.edu/}},
  note={Accessed: 2025-01-10}
}
